BMC Medical Informatics and
Decision Making

**Open Access**

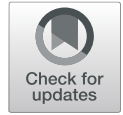

# Two-stage framework for optic disc localization and glaucoma classification in retinal fundus images using deep learning

Muhammad Naseer Bajwa[1,2*] , Muhammad Imran Malik[3,4], Shoaib Ahmed Siddiqui[1,2], Andreas Dengel[1,2], Faisal Shafait[3,4], Wolfgang Neumeier[5] and Sheraz Ahmed[2]

## Abstract

**Background:** With the advancement of powerful image processing and machine learning techniques, Computer Aided Diagnosis has become ever more prevalent in all fields of medicine including ophthalmology. These methods continue to provide reliable and standardized large scale screening of various image modalities to assist clinicians in identifying diseases. Since optic disc is the most important part of retinal fundus image for glaucoma detection, this paper proposes a two-stage framework that first detects and localizes optic disc and then classifies it into healthy or glaucomatous.

**Methods:** The first stage is based on Regions with Convolutional Neural Network (RCNN) and is responsible for localizing and extracting optic disc from a retinal fundus image while the second stage uses Deep Convolutional Neural Network to classify the extracted disc into healthy or glaucomatous. Unfortunately, none of the publicly available retinal fundus image datasets provides any bounding box ground truth required for disc localization. Therefore, in addition to the proposed solution, we also developed a rule-based semi-automatic ground truth generation method that provides necessary annotations for training RCNN based model for automated disc localization.

**Results:** The proposed method is evaluated on seven publicly available datasets for disc localization and on ORIGA dataset, which is the largest publicly available dataset with healthy and glaucoma labels, for glaucoma classification. The results of automatic localization mark new state-of-the-art on six datasets with accuracy reaching 100% on four of them. For glaucoma classification we achieved Area Under the Receiver Operating Characteristic Curve equal to 0.874 which is 2.7% relative improvement over the state-of-the-art results previously obtained for classification on ORIGA dataset.

**Conclusion:** Once trained on carefully annotated data, Deep Learning based methods for optic disc detection and localization are not only robust, accurate and fully automated but also eliminates the need for dataset-dependent heuristic algorithms. Our empirical evaluation of glaucoma classification on ORIGA reveals that reporting only Area Under the Curve, for datasets with class imbalance and without pre-defined train and test splits, does not portray true picture of the classifier's performance and calls for additional performance metrics to substantiate the results.

* Correspondence: bajwa@dfki.uni-kl.de
[1]Fachbereich Informatik, Technische Universität Kaiserslautern, 67663
Kaiserslautern, Germany
[2]Deutsche Forschungszentrum für Künstliche Intelligenz GmbH (DFKI), 67663
Kaiserslautern, Germany
Full list of author information is available at the end of the article

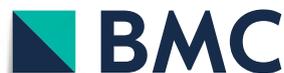

## Background

Glaucoma is a syndrome of eye disease that leads to subtle, gradual, and eventually total loss of vision if untreated. The disordered physiological processes associated with this disease are multifactorial. However, the causes of glaucoma are usually associated with the build-up of IntraOcular Pressure (IOP) in the eye that results from blockage of intraocular fluid drainage [1]. Although the exact cause of this blockage is unknown, it tends to be inherited and can be linked to old age, ethnicity, steroid medication, and other diseases like diabetes [2]. The increased IOP damages the optic nerve that carries visual information of photo receptors from eye to brain. Generally glaucoma does not show any signs or symptoms until it has progressed to advanced stage at which point the damage becomes irreversible. It has been reported that the damage to optic nerve fibres becomes noticeable and reduction in visual field is detected when about 40% of axons are already lost [3]. However, it is possible to slow down the impairment caused by glaucoma if it is diagnosed sufficiently early. Recently, World Health Organization (WHO) recognized glaucoma as the third biggest cause of blindness after un-operated cataract and uncorrected refractive errors [4] and the leading cause of irreversible vision loss.

Glaucoma is normally diagnosed by obtaining medical history of patients, measuring IOP, performing visual field loss test, and conducting manual assessment of Optic Disc (OD) using ophthalmoscopy to examine the shape and colour of optic nerve [5, 6]. Optic Disc is the cross sectional view of optic nerve connecting to the retina of each eye. It looks like a bright round spot in retinal fundus image. In case of glaucoma, the IOP damages the nerve fibres constituting optic nerve. As a result OD begins to form a cavity and develops a crater-like appearance, at the front of the nerve head, called Optic Cup (OC). The boundary of the disc also dilates and the colour changes from healthy pink to pale. The Cup-to-Disc Ratio (CDR) is one of the major structural image cues considered for glaucoma detection [7]. Figure 1 shows healthy optic disc and its condition during various stages of glaucoma.

In retinal images, some of the important structural indications of glaucoma are disc size, CDR, Ratio of Neuroretinal Rim in Inferior, Superior, Nasal and Temporal quadrants (ISNT rule), and Peripapillary Atrophy (PPA) [9] etc. These indications are usually concentrated in and around OD. Therefore, segmentation of this Region Of Interest (ROI), that is detecting the contour of OD, is not only useful for a more focused clinical assessment by the ophthalmologists but also helpful in training a computer based automated method for classification. However, automated glaucoma detection techniques based upon segmented discs are very sensitive to the accuracy of segmentation and even a small error in delineation of OD may affect the diagnosis significantly [10]. Localization, on the other hand, gives the exact location of OD in the whole image with some surrounding context. Automatic methods for glaucoma detection based upon this approach of ROI extraction are more resilient to localization errors.

From automated classification point of view the disease pattern in retinal fundus images is inconspicuous and complex. Detecting ROI from natural scene images is comparatively easy because it has an obvious visual appearance, for example colour, shape and texture etc. In contrast, the significant features of disease in medical images are hidden and not readily discernible except by highly trained and qualified field experts. Deep Learning, however, has been shown to learn discriminative representation of data that can identify otherwise unnoticeable characteristics. Such algorithms achieve this useful and compact representation of data by applying multiple linear and non-linear transformations on training data [11] in a cascaded fashion. Such Computer Aided Diagnosis (CAD) can be very helpful in providing standardized and cost effective screening at a larger scale. These automated systems may reduce human error, offer timely services at remote areas, and are free from clinician's bias and fatigue. In this work we address localization and classification of OD in colour retinal fundus images. Figure 2 shows our proposed two-stage system. The framework is able to surpass current state-of-the-art in both localization as well as classification tasks. The contributions of this work are listed below.

- To the best of our knowledge there is no fully automated disc localization method that can give robust and accurate results independent of the datasets. Also many existing heuristic methods, for example [12–14], set the bar for correct localization as low as accepting a predicted disc location correct if Intersection Over Union (IOU) between actual and predicted locations is greater than 0. To address these issues we propose a dataset-independent fully automated disc localization method based on faster RCNN (Regions with Convolutional Neural Network) as shown in Fig. 2a. Our approach sets new state-of-the-art on six out of seven datasets for localization while setting the bar for correct localization at IOU > 50.
- We used Deep Convolutional Neural Network (DCNN) as shown in Fig. 2b on ODs extracted in stage one to classify the images into healthy and glaucoma affected images. Our classification results surpass previous best method with 2.7% relative improvement in terms of Area Under the Curve (AUC) of Receiver Operating Characteristic (ROC) curve.

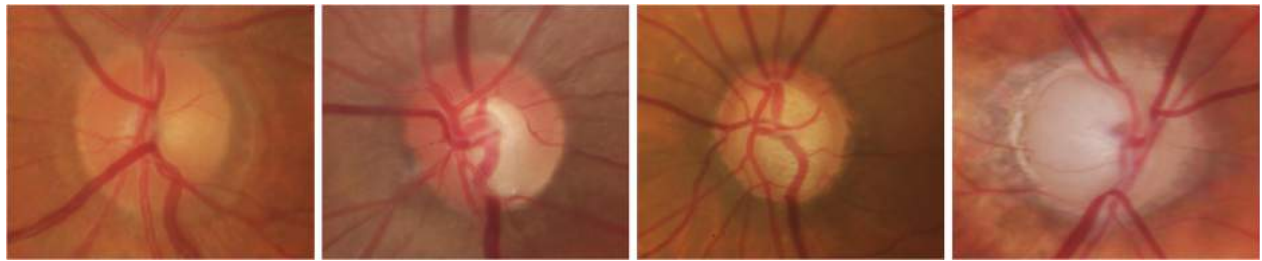

**Fig. 1** Stages of glaucoma in retinal fundus images taken from Rim-One dataset [8]. **a** Healthy Disc **b** Early Glaucoma **c** Moderate Glaucoma **d** Severe Glaucoma

- Training faster RCNN based methods requires bounding box annotations for desired objects. This ground truth was not available with any publicly available dataset used. Therefore, we also developed our own semi-automated ground truth generation method using a rule-based algorithm.
- Our empirical evaluation demonstrates that reporting only Area Under the Curve (AUC) for datasets with class imbalance and without standard formulation of train and test splits, as is the case with ORIGA, does not portray true picture of the classifier's performance and calls for additional performance metrics to corroborate the results. Therefore, we provide a detailed report of our classification results using precision, recall, and f-scores in addition to AUC for fair analysis and thorough comparison with other methods in the future.

The rest of this paper is organized as follows. Rest of this section gives a brief review of the state-of-the-art approaches for OD localization and glaucoma detection. The section on Methods provides comprehensive details of the whole methodology starting from an introduction to seven datasets used in this work, establishing the rationale for and explaining our semi-automated ground truth generation mechanism and finally presenting the proposed two-stage solution for fully automated localization and classification using deep learning. Results and Discussion section highlights the significance of the findings and finally the last section concludes the discussion with perspective extension of this work in the future.

## Related work

Early diagnosis of glaucoma is vital for timely treatment of patients. Medical practitioners have proposed a number of criteria for early diagnosis and these criteria mostly focus on or around OD region. If the position, centre, and size of OD is calculated accurately it can greatly help in further automated analysis of the image modality. Rest of this subsection discusses various image processing and machine learning approaches making use of these diagnosis criteria for disc localization and glaucoma identification.

### Localization of optic disc

Although optic disc can be spotted manually as a round bright spot in a retinal image, yet performing large scale manual screening can prove to be really tiresome, time consuming, and prone to human fatigue and predisposition. CAD can provide efficient and reliable alternative solution with near human accuracy (as shown in Table 4). Usually the disc is the brightest region in the image. However, if ambient light finds its way into the image while capturing the photo it can look brighter than optic disc. Furthermore, occasionally some shiny reflective areas appear in the fundus image during image capturing. These shiny reflections can also look very bright and mislead a heuristic algorithm in considering them as candidate regions of interest. There are many approaches laid out by researchers for OD localization exploiting different image characteristics. Some of these approaches are briefly covered below.

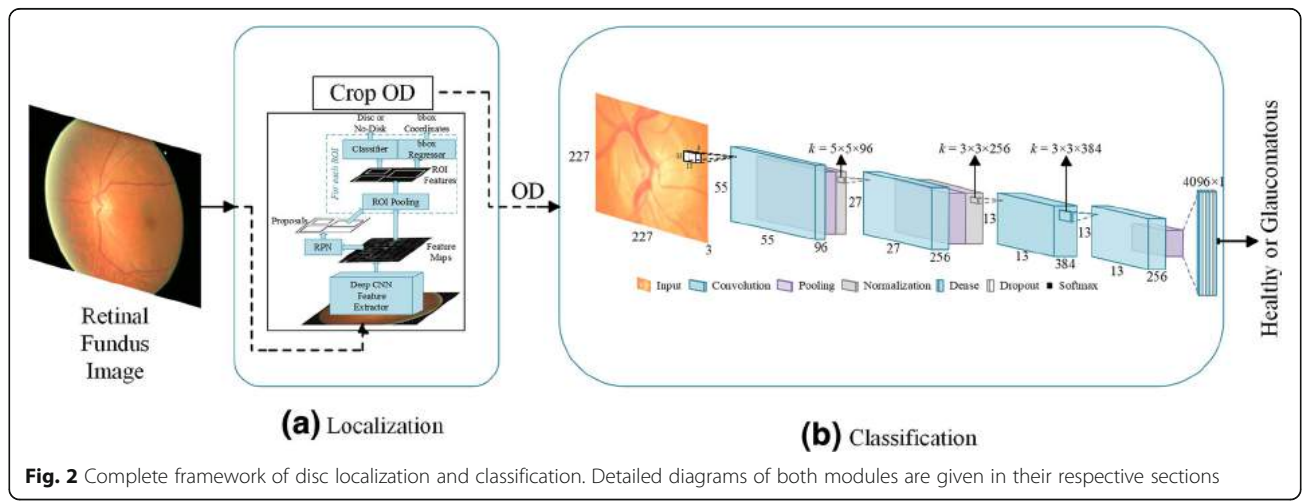

**Fig. 2** Complete framework of disc localization and classification. Detailed diagrams of both modules are given in their respective sections

Intensity variations in the image can help locate optic disc in fundus images. To make use of this variation the image contrast is first improved using some locally adaptive transforms. The appearance of OD is then identified by noticing rapid variation in intensity as the disc has dark blood vessels alongside bright nerve fibres. The image is normalized and average intensity variance is calculated within a window of size roughly equal to expected disc size. The disc centre is marked at the point where the highest intensity is found. Eswaran et al. [12] used such intensity variation based approach. They applied a $25 \times 35$ averaging filter with equal weights of 1 on the image to smooth it and get rid of low intensity variations and preserve ROI. Chràstek et al. [13] used $31 \times 31$ averaging filter and the ROI is assumed to be $130 \times 130$ pixels. They used Canny Edge Detector [14] to plot the edges in the image. To localize the optic disc region they used only green channel of RGB image. Abràmoff et al. [15] proposed that the optic disc can be selected by taking only top 5% brightest pixels and hue values in the yellow range. The surrounding pixels are then clustered to constitute a candidate region. The clusters which are below a certain threshold are discarded. Liu et al. [16] used a similar approach. They first divided the image into $8 \times 8$ pixels grid and selected the block with maximum number of top 5% brightest pixels as the centre of the disc. Nyúl [17] employed an adaptive thresholding with a window whose size is determined to approximately match the size of the vessel thickness. A mean filter with the large kernel is then used with threshold probing for rough localization.

Another extensively used approach is threshold based localization. A quick look at the retinal image tells that the optic disc is mostly the brightest region in the image. This observation is made and exploited by many including Siddalingaswamy and Prabhu [18]. It is also noticed that the green channel of RGB has the greatest contrast compared to red and blue channels [19–21], however, red channel has also been used [22] due to the fact that it has fewer blood vessels that can confuse the rule-based localization algorithm. Optimal threshold is chosen based upon approximation of image histogram. The histogram of the image is gradually scanned from a high intensity value $I_1$, slowly decreasing the intensity until it reaches a lower value $I_2$ that produces at least 1000 pixels with the same intensity. It results in a subset of histogram. The optimal threshold is taken as the mean of the two intensities $I_1$ and $I_2$. Applying this threshold produces a number of connected candidate regions. The region with the highest number of pixels is taken as the optic disc. Dashtbozorg et al. [23] used Sliding Band Filter (SBF) [24] on downsampled versions of high resolution images since SBF is computationally very expensive. They apply this SBF first to a larger region of interest on downsampled image to get a rough localization. The position of this roughly estimated ROI is then used to establish a smaller ROI on original sized image for a second application of SBF. The maximum filter response results in $k$-candidates pointing to potential OD regions. They then use a regression algorithm to smooth the disc boundary. Zhang et al. [25] proposed a fast method to detect optic disc. Three vessel distribution features are used to calculate possible horizontal coordinates of the disc. These features are local vessel density, compactness of the vessels and their uniformity. The vertical coordinates of the disc are calculated using Hough Transform according to the global vessel direction characteristics.

Hough Transform (HT) has been widely utilized to detect OD [25–27] due to disc's inherent circular shape and bright intensity. The technique is applied to binary images after they have undergone morphological operations to remove noise or reflection of light from ocular fundus that may interfere with the calculation of Hough Circles. The HT maps any point $(x, y)$ in the image to a circle in a parameter space that is characterized by centre $(a, b)$ and radius $r$, and passes through the point $(x, y)$ by following the equation of circle. Consequently, the set of all feature points in the binary image are associated with circles that may almost be concentric around a circular shape in the image for some given value of radius $r$. This value of $r$ should be known a priori by experience or experiments. Akyol et al. [28] presented an automatic method to localize OD from retinal images. They employ keypoint detectors to extract discriminative information about the image and Structural Similarity (SSIM) index for textual analysis. They then used visual dictionary and random forest classifier [29] to detect the disc location.

### Glaucoma classification

Automatic detection and classification of glaucoma has also been widely studied by researcher since long. A brief overview of some of the current works is presented below. For a thorough coverage of glaucoma detection techniques [30–32] may be consulted.

Fuente-Arriaga et al. [33] proposed measuring blood vessels displacement within the disc for glaucoma detection. They first segment vascular bundle in OD to set a reference point in the temporal side of the cup. Centroid positions of inferior, superior, and nasal vascular bundles are then determined which are used to calculate $L1$ distance between centroid and normal position of vascular bundles. They applied their method on a set of 67 images carefully selected for clarity and quality of retina from a private dataset and report 91.34% overall accuracy. Ahmad et al. [34] and Khan et al. [35] have used almost similar techniques to detect glaucoma. They calculate CDR and ISNT quadrants and classify an image as glaucomatous if the CDR is greater than 0.5 and it violates ISNT rule. Ahmad et al. applied the method on 80 images taken from DMED dataset, FAU data library, and Messidor dataset and reported 97.5% accuracy whereas Khan et al. used 50 images taken from the above-mentioned datasets and reported 94% accuracy. Though the accuracies reported by the aforementioned researchers are well above 90%, their test images are handpicked and so fewer in number that the results are not statistically significant and cannot be reliably generalized to large scale public datasets.

ORIGA [36] is a publicly available dataset of 650 retinal fundus images for benchmarking computer aided segmentation and classification. Xu et al. [37] formulated a reconstruction based method for localizing and classifying optic discs. They generate a codebook by random sampling from manually labelled images. This codebook is then used to calculate OD parameters based on their similarity to the input and their contribution towards reconstruction of input image. They report AUC for glaucoma diagnosis at 0.823. Noting that classification based approaches perform better than segmentation based approaches for glaucoma detection, Li et al. [38] proposed to integrate local features with holistic feature to improve glaucoma classification. They ran various CNNs like AlexNet, VGG-16 and VGG-19 [39] and found that combining holistic and local features with AlexNet as classifier gives highest AUC at 0.8384 using 10-fold cross validation, while the manual classification gives AUC equal to 0.8390 on ORIGA dataset. Chen et al. [6] also used deep convolutional networks based approach for glaucoma classification on ORIGA dataset. Their method inserts micro neural networks within more complex models so that the receptive field has more abstract representation of data. They also make use of a contextualization network to get hierarchal and discriminative representation of images. Their achieved AUC is 0.838 with 99 randomly selected train images and rest for testing. In another of their publications Chen et al. [5] used a six layer CNN to detect glaucoma from ORIGA images. They used the same strategy of taking 99 random images for training and rest for testing and obtained AUC at 0.831.

Recently, Al-Bander et al. [40] used deep learning approach to segment optic cup and OD from fundus images. Their segmentation model has a U-Shape architecture inspired from U-Net [41] with Densely connected convolutional blocks, inspired from DenseNet [42]. They outperformed state-of-the-art segmentation results on various fundus datasets including ORIGA. For glaucoma diagnosis, however, in spite of combining commonly used vertical CDR with horizontal CDR, they were able to achieve AUC at 0.778 only. Similarly Fu et al. [43] also proposed a U-Net like architecture for joint segmentation of optic cup and OD and named it M-Net. They added a multi-scale input layer that gets the input image at various scales and gives receptive fields of respective sizes. The main U-shaped convolutional network learns hierarchical representation. The so-called side-output layers generate prediction maps for early layers. These side-output layers not only relieve vanishing gradient problem by back propagating side-output loss directly to the early layers but also help achieve better output by supervising the output maps of each scale. For glaucoma screening on ORIGA data set, they trained their model on 325 images and tested on rest of 325 images. Using vertical CDR of their segmented discs and cups they achieved AUC at 0.851.

## Methods

This section presents the whole methodology of optic disc localization and classification starting from a brief introduction of some of the publicly available retinal fundus image datasets that have been used in this work. It can be noticed from this introduction that none of these datasets provided any bounding box ground truth for disc localization and, therefore, prompting for development of ground truth generation mechanism.

### Datasets used in this work

#### ORIGA (−light)

ORIGA [36] (an Online Retinal fundus Image database for Glaucoma Analysis and research) aims to provide clinical ground truth to benchmark segmentation and classification algorithms. It uses a custom developed tool to generate manual segmentation for OD and OC. It also provides CDR and labels for each image as glaucomatous or healthy. This dataset has been used as a standard dataset in some of the recent state-of-the-art researches for glaucoma classification. The dataset was collected by Singapore Eye Research Institute and has 482 healthy images and 168 glaucomatous images.

#### HRF image database

High Resolution Fundus [44] (HRF) Image database is provided by Department of Ophthalmology, Friedrich-

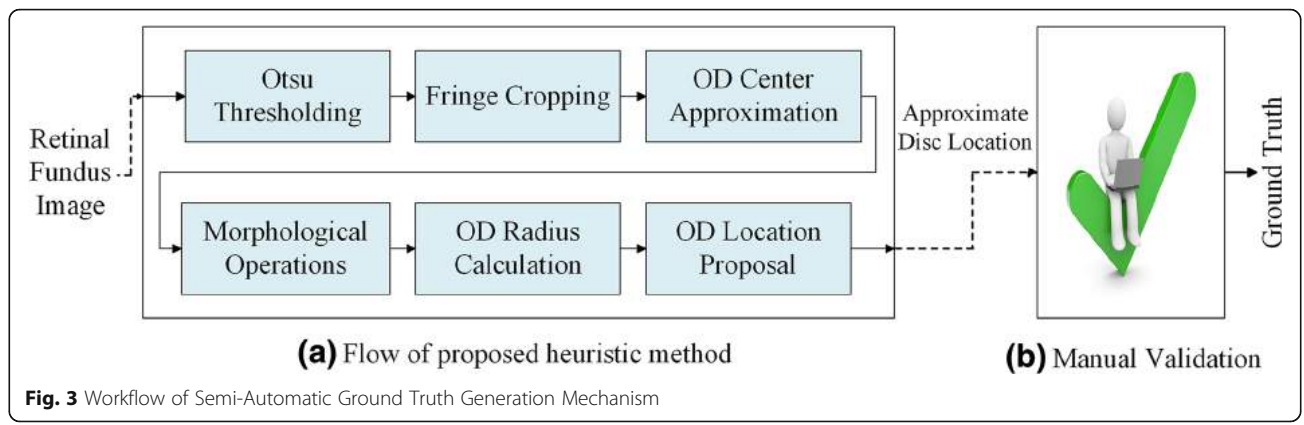

**(a)** Flow of proposed heuristic method    **(b)** Manual Validation

**Fig. 3** Workflow of Semi-Automatic Ground Truth Generation Mechanism

Alexander University Erlangen-Nuremberg, Germany. It consists of 15 healthy images, 15 glaucomatous image sand 15 images with diabetic retinopathy. For each image, binary gold standard vessel segmentation is provided by a group of experts and clinicians.

### OCT & CFI

This dataset [45] contains Optical Coherence Tomography (OCT) and Colour Fundus Images of both eyes of 50 healthy persons collected at Ophthalmology Department of Feiz Hospital, Isfahan, Iran. As the images were taken as part of a study on comparison of macular OCTs in right and left eyes of normal people, it doesn't provide any ground truth with respect to segmentation of OD or blood vessels, or OD localization.

### DIARETDB1

Standard DIAbetic RETinopathy DataBase calibration level 1 (DIARETDB1) [46] is publicly available dataset consisting of 89 colour retinal fundus images taken at Kuopio University Hospital, Finland. The prime objective of this dataset is to benchmark the performance of automated methods for diabetic retinopathy detection. Four independent medical expert are employed to annotate the dataset and provide the markings for microaneurysms, haemorrhages, and hard and soft exudates. Based upon the markings provided, 84 of the images were found to have at least mild non-proliferative diabetic retinopathy while rest of five images were found to be healthy. The dataset does not provide retinopathy grades in accordance with International Clinical Diabetic Retinopathy (ICDR) severity grade or ground truth for OD localization.

### DRIVE

Digital Retinal Images for Vessel Extraction (DRIVE) [47] consists of 40 images taken in The Netherlands as part of a diabetic retinopathy screening programme. The dataset is divided into train and test splits. Train set contains 20 images with single manual segmentation mask for blood vessels. Test set also contains 20 images with two manual segmentation masks. This dataset also does not provide any annotation for optic disc localization.

### DRIONS-DB

Digital Retinal Images for Optic Nerve Segmentation DataBase [48] commonly known as DRIONS-DB is dataset for benchmarking Optic Nerve Head (ONH) segmentation from retinal images. The data were collected at Ophthalmology Service at Miguel Servet Hospital, Saragossa, Spain and contains 110 images. It provides ONH contours traced by two independent experts using a software tool for image annotation.

### Messidor

Methods to evaluate segmentation and indexing techniques in the field of retinal ophthalmology (Messidor) [49] is a large publicly available dataset of 1200 high resolution colour fundus images. The dataset contains 400 images collected from three ophthalmologic departments each, under a project funded by French Ministry of Research and Defense. It provides diabetic retinopathy grade for each image from 0 (healthy) to 3 (severe) as well as risk of macular edema at a scale of 0 (no risk) to 2 (high risk).

This research was started with the aim to classify retinal fundus images into healthy and glaucomatous.

**Table 1** Overview of datasets used for evaluating heuristic method

| Dataset | Total Size | Healthy | Glaucoma | Split | | |
|---|---|---|---|---|---|---|
| | | | | Train | Validate | Test |
| ORIGA | 650 | 482 | 168 | 441 | 36 | 173 |
| HRF | 30 | 15 | 15 | 12 | 04 | 14 |
| OCT & CFI | 100 | 100 | Nil | 72 | 20 | 08 |
| Total | 780 | 597 | 183 | 525 | 48 | 207 |

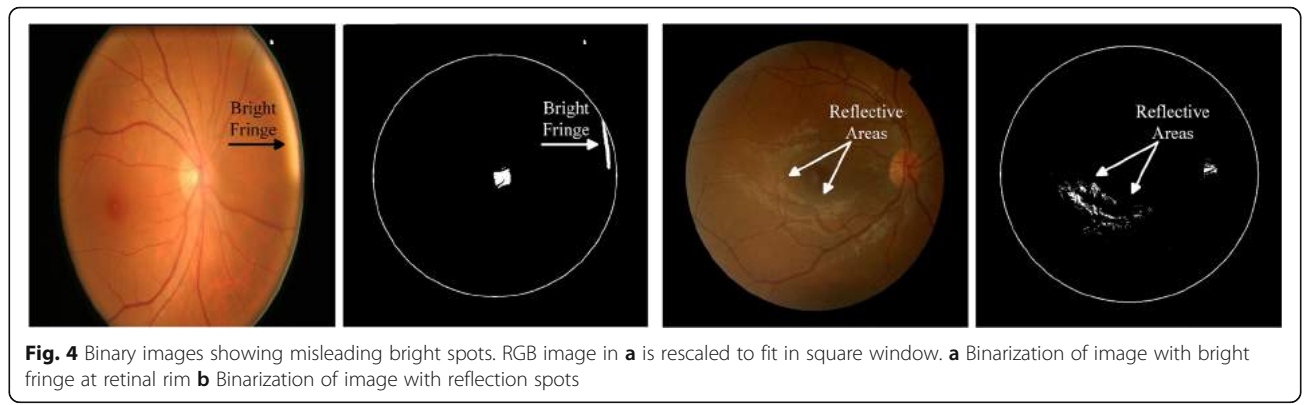

**Fig. 4** Binary images showing misleading bright spots. RGB image in **a** is rescaled to fit in square window. **a** Binarization of image with bright fringe at retinal rim **b** Binarization of image with reflection spots

However, since the very beginning we learned from both experiments and literature that feeding the whole retinal fundus image to CNN does not help the network focus on the most important part of the image, i.e. the OD, for the given classification task. Therefore, it was imperative to extract the ROI and feed it to the network. A quick survey on disc localization methods revealed that the existing methods were heuristic and developed specifically for one or a few datasets. These methods could not generalize well on other datasets. There was a serious need for a fully automated, robust and dataset independent method for disc localization that can be applied to a vast number of retinal fundus image datasets with reliability and high accuracy. This was possible using supervised neural network approach. However, this approach required availability of ground truth for training, which was not found with any of the seven datasets used in this work. Therefore, it was deemed fit to first devise a method for ground truth generation with minimal human involvement. The generated ground truth would then be used to train and benchmark the performance of fully automated disc localization method. In the following sub-sections, step by step progress from GT generation to disc extraction and finally classification is discussed in detail.

## Semi-automated GT generation for OD localization

We developed a heuristic method to approximate the location of OD in retinal images. Results generated by this heuristic method are manually checked and necessary corrections are made where needed. Figure 3 depicts the workflow of this mechanism. It consists of a heuristic algorithm that gives a proposal for OD location which is then manually verified by an expert. This way we generated localization ground truth for all seven datasets discussed in the previous section.

Three publicly available datasets of high resolution colour retinal fundus images were chosen to evaluate the performance of heuristic localization algorithm. Table 1 gives an overview of the datasets used. Out of 780 images, 525 were randomly selected for training, 48 images were taken for validation and the rest of 207 images were kept aside for testing. The validation set was used to find various empirical parameters like retinal rim crop margin and maximum size of valid disc radius etc. These empirical parameters were selected manually such that they gave highest validation accuracy. Once these values are calculated, they were fixed during testing. To ensure that these values work on images from different datasets having different resolutions, we rescaled all images to a fixed size (1500 × 1500), processed the images, and then rescaled them back to their original size. The mixture of three different datasets introduces enough inter-dataset variations in the images to thoroughly and rigorously validate the accuracy and robustness of the heuristic method.

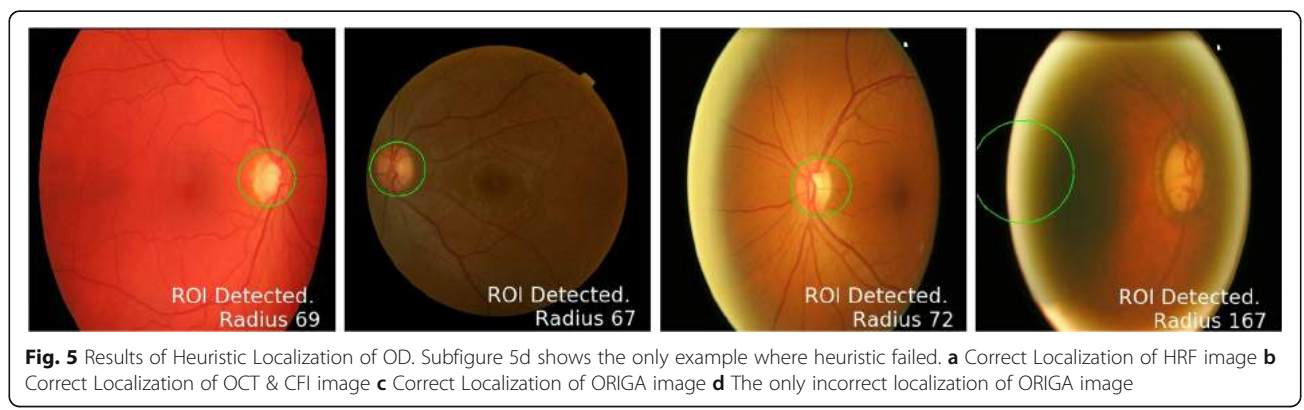

**Fig. 5** Results of Heuristic Localization of OD. Subfigure 5d shows the only example where heuristic failed. **a** Correct Localization of HRF image **b** Correct Localization of OCT & CFI image **c** Correct Localization of ORIGA image **d** The only incorrect localization of ORIGA image

**Table 2** IOU of heuristic predictions and ground truth

| IOU (%) | 20 | 50 | 60 | 70 | 80 |
|---|---|---|---|---|---|
| Test Accuracy | 99.52 | 96.14 | 75.96 | 51.97 | 09.18 |

*Heuristic algorithm for OD localization*
This section details our algorithm to find approximate OD location from retinal images. The basic flow of the method is shown in Fig. 3a. It can be observed from the data that OD is usually the brightest region in retinal fundus image. However, there could be other bright spots in the image, due to some disease or imperfect image capturing conditions that can affect the performance of any empirical or heuristic method. Figure 4 shows two examples of such misleading bright spots.

The first column of each sub figure shows colour retinal fundus image and the second column shows the binary image corresponding to the respective colour image. The bright fringe at the retinal rim, as shown in Fig. 4a, occurs when a patient does not place his/her eye correctly on the image capturing equipment and the ambient lights gets through the corners of the eye. Figure 4b shows example of shiny cloud like spots around macular region caused by reflection of light from ocular fundus which is a common phenomenon in younger patients.

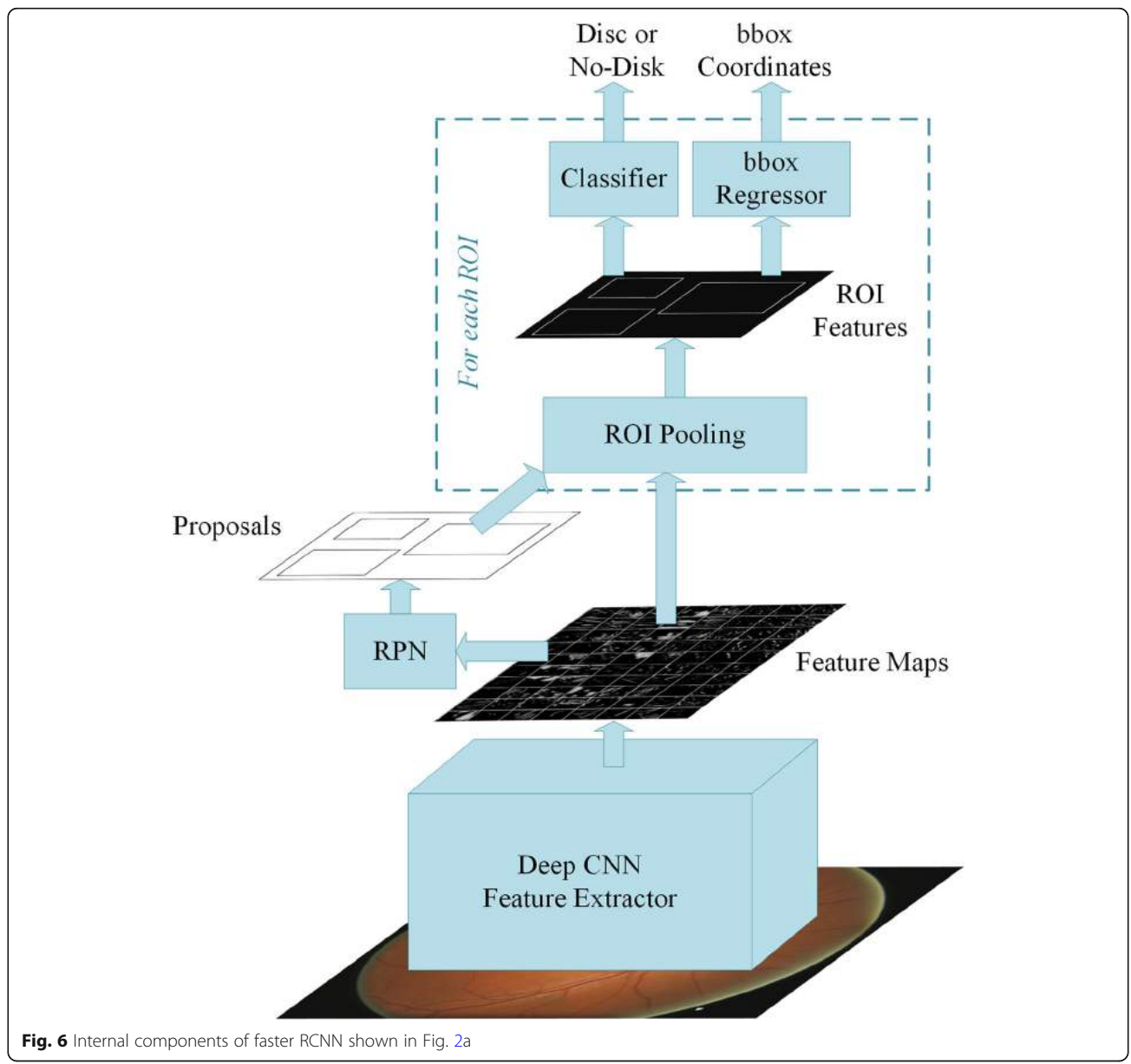

**Fig. 6** Internal components of faster RCNN shown in Fig. 2a

**Table 3** Comparison of IOU of heuristic and automated methods

| IOU (%) | 20 | 50 | 60 | 70 | 80 |
|---|---|---|---|---|---|
| Heuristic Method | 99.52 | 96.14 | 75.96 | 51.97 | 09.18 |
| Automated Method | 100.0 | 100.0 | 100.0 | 99.52 | 94.69 |

In our heuristic method the fringe is removed by first finding the diameter of the retinal rim inside the image. This is done by applying Otsu thresholding [50] on the image. Otsu binarization method assumes that the image consists of only two classes of pixels (foreground and background) following a bi-model histogram. It adaptively calculates the most appropriate threshold value that can categorize all the pixels into two classes. As a result it largely turns the retina into a white disc and keeps the background black. This output is used to calculate the centre and radius of the retina. A circular mask with radius less than retinal radius is created and applied to the original image to crop and possibly get rid of the fringe. Although Otsu lacks the precision to accurately segment the fundus area but it gives a quick and automated way to calculate approximate fundus dimensions for cropping unwanted artefacts on the rim.

A custom adaptive binarization is then applied on the resultant image with threshold for each image calculated as the mean of top 1% brightest pixels. This technique locates approximate core of OD. Before finding the centre of this approximate OD, we apply morphological erosion operation to remove small reflective areas and random impulse noise. This is followed by dilation operation to connect disjoint white spots into fewer and bigger connected blobs. The result of these operations is a better approximation of OD. Radius and centre of this approximate disc location is then calculated and a circle with radius greater than calculated radius is drawn on the image to identify and localize OD. Lastly, these proposed locations are manually verified by expert and necessary corrections are made where necessary.

### Results of heuristic localization

Visual inspection of output of train and test datasets showed that the method failed on only 3 out of 573 (test + validate) images and on only 1 of 207 test images from three different datasets as shown in Fig. 5. To quantify the accuracy of this approach we calculated IOU between bounding boxes given by proposed method and manual ground truth. Table 2 shows the accuracy of the this method in terms of overlap between predicted disc and actual disc.

The results show that more than 96% of ODs are localized with more than 50% of actual disc present in the prediction. Also, about 52% of the predicted discs contain more than 70% of the actual disc. The average overlap between predicted disc area and ground truth for the test images is around 70%. It is also worth mentioning here that the minimum IOU of a correctly localized disc in this method is more than 20% whereas some researchers [51–53] have opted to consider their localization correct if the distance between predicted disc centre and actual disc centre is less than expected disc diameter — in other words if IOU > 0.

Although the results of heuristic based approach are very promising, yet they are dataset specific and might not work well in real world scenarios on a diverse spectrum of fundus images. Therefore, in next section we explore a fully automated approach of precise disc localization without using any empirical knowledge about the dataset. Necessary corrections are made in the annotations given by heuristic approach and these semi-automated annotations were provided to automated localization method as ground truth.

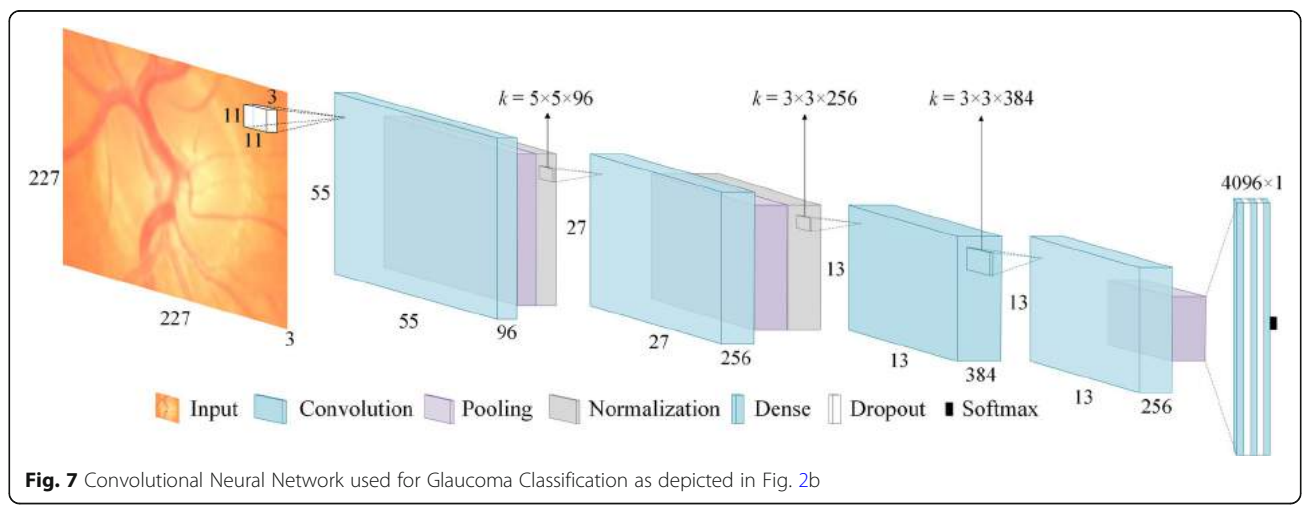

**Fig. 7** Convolutional Neural Network used for Glaucoma Classification as depicted in Fig. 2b

## Two-stage method for automated disc localization and classification: the proposed approach

Having generated necessary GT for the training of automated method, this section now details the complete solution for reliable and robust OD localization and its classification into healthy or glaucomatous.

### Automated disc localization

Existing approaches for disc localization are dataset dependent and make use of empirical knowledge of the datasets. For our two-stage solution to work, we adopted fully automated localization using faster RCNN [54], which is a unified network for object detection. As shown in Fig. 6, the model consists of three major modules: Region Proposal Network (RPN), CNN classifier, and Bounding Box Regression. Given an image for object detection RPN generates a number of random rectangular object proposals with associated *objectness* scores. These proposals are fed to CNN that classifies whether a given object is present in the proposal. Then bounding box regression is performed to fit the rectangle closely to the object and provide precise location of the object in the image. The disc localization outcome of faster RCNN on three datasets, shown in Table 1, is given in Table 3. As can be seen in the table, faster RCNN gives 100% correct localization for 60% IOU and average overlap of 97.11% on these three datasets combined. Results of automated localization on other datasets and detailed analysis thereof can be found in the results section.

### Classification of glaucoma

In the first stage, we extracted OD because most of the glaucoma related information is contained in this region [5, 6]. Extracting this ROI not only produces a smaller initial image that is computationally efficient but also allows deep neural network to focus on the most important part of the image. Figure 7 depicts the architecture of the CNN used in this work.

The network consists of four convolutional layers followed by three fully connected layers. Max pooling with overlapping strides and local response normalization are used after first two convolutional layers. Max Pooling also follows fourth convolutional layer. First two fully connected layers are followed by dropout layers with dropout probability of 0.5. The output of last dense layer is fed to softmax function that gives prediction probabilities for each class. A brief overview of the network architecture is given below.

### Convolutional layers

From a high level point of view, convolutional layers are used to learn important features from an image. The layers in the beginning of the network tend to learn simple features like curves, lines and edges whereas subsequent layers learn more complex and abstract features such as hands and ears. Mathematically, this is done by first taking a feature detector or kernel of size $k \times k$, sliding it on the input space and performing convolution between kernel and input patch at each point. The size of the kernel is usually smaller than input space. The depth of the kernel, however, has to be the same as the input depth. Multiple kernels are used on each convolutional layer to better preserve the spatial dimensions. Each kernel looks for a specific feature in the input space and produces a feature map. In the network shown in Fig. 7, the first convolutional layer uses 96 kernels each of size $11 \times 11 \times 3$. As the convolution is a linear operation (element wise multiplication of kernel and input patch values followed by their sum), performing multiple convolutions in multiple layers ends up in one big linear operation and, therefore, limit the learning ability of the network. To address this problem the output of each convolutional layer is passed through a non-linear function. Rectified Linear Unit (ReLU), defined as $f(x) = max\ (0, x)$, is the most popular nonlinear function used to increase the nonlinear properties of the network.

### Pooling layers

Pooling layers are used to downsample the feature maps without losing significant information. This is done by taking a small window of size $p \times p$ from each slice of feature map and giving, for example, average value of that window. The window is then slid over the whole feature map with stride of $s$ pixels. When $s < p$ we get overlapping pooling. It not only reduces the size of the feature map but also helps in controlling the overfitting which is a major concern in deep networks with small datasets. In our network we have used MaxPooling (picking the maximum value from the window) with window size $3 \times 3$ and stride of 2 pixels.

**Table 4** Performance of automated disc localization algorithm on unseen datasets

| Algorithms | Criterion (IOU >) | DIARETDB1 $N = 89$ | DRIVE $N = 40$ | DRIONS-DB $N = 110$ | Messidor $N = 1200$ |
|---|---|---|---|---|---|
| Our Method (RCNN-based) | 50 | 100.0 | 97.50 | 99.09 | 99.17 |
| Giachetti et al. [12] | 0 | N/A | N/A | N/A | 99.83 |
| Yu et al. [13] | 0 | N/A | N/A | N/A | 99.08 |
| Aquino et al. [14] | 0 | N/A | N/A | N/A | 98.83 |
| Akyol et al. [31] | 50 | 94.38 | 95.00 | N/A | N/A |
| Qureshi et al. [56] | 50 | 94.02 | 100.0 | N/A | N/A |
| Godse et al. [57] | 50 | 96.62 | 100.0 | N/A | N/A |
| Lu et al. [58] | 50 | 96.91 | N/A | N/A | N/A |

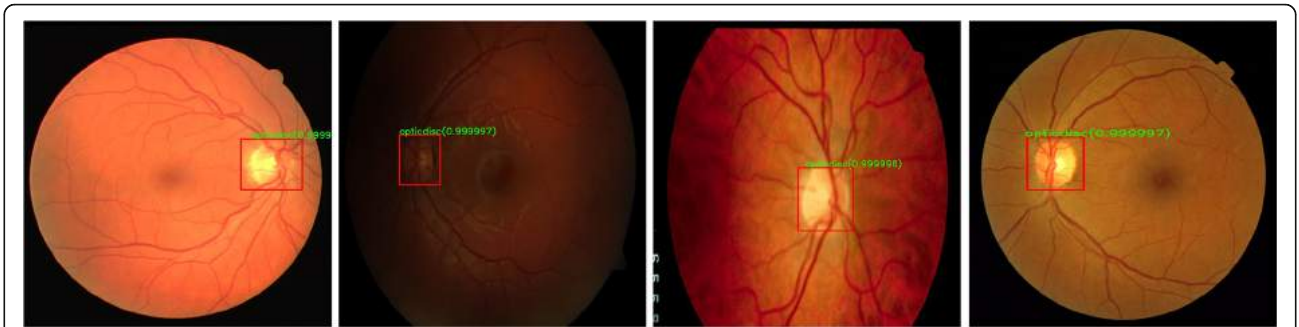

**Fig. 8** Results of Automated Localization on different datasets. Notice the illumination and contrast variations amongst the datasets. **a** Sample image from DRIVE **b** Sample image from DIARETDB1 **c** Sample image from DRIONS-DB **d** Sample image from Messidor

### Normalization layers

A Local Response Normalization (LRN) layer is the mathematical equivalent of neurobiological lateral inhibition which is the ability of an excited neuron to dominate its neighbours. As the activation of ReLU might get unbounded for some neurons, the LRN is used to normalize it around the local neighbourhood of such excited neurons. This creates a competition for big activities among neuron outputs computed by various kernels and also helps in better generalization.

### Dropout layers

Dropout layers are also helpful in preventing overfitting and aiding generalization during training. This layer is implemented by setting the outputs of hidden neurons equal to zero with a given probability. These dropped out neuron thus don't contribute in the current training pass. This way the network is mandated to learn more robust features. The network of Fig. 7 uses dropout probability equal to 0.5.

**Table 5** Detailed performance measures of CNN classifier using random training

| Class | Precision (%) | Recall (%) | F1-Score | No. of Images |
|---|---|---|---|---|
| Healthy | 81.12 | 94.9 | 0.8747 | 412 |
| Glaucoma | 69.57 | 34.53 | 0.4615 | 139 |
| Total | 78.21 | 79.67 | 0.7705 | 551 |

## Results and discussion

This section reports and analyses the results of localization and classification steps detailed in previous section.

### Results of automated disc localization

For automated localization of OD the model was trained for 100,000 iterations using VGG16 as classifier pre-trained on Pascal VOC2007 [55]. The GT generated by our semi-automated method is used along with 573 images, previously employed for training and validation of heuristic method, to train the network. The disc localization outcome of the automated method is given in Table 3.

Once trained and evaluated on ORIGA, HRF and OCT& CFI datasets, the model was also tested on other publicly available databases and the results are compared with some state-of-the-art methods developed specifically for those datasets. The results highlight the comparative performance of fully automated method with state-of-the-art heuristic algorithms. The accuracies of our method are taken for 50% IOU. The results reported by [51–53] are for IOU > 0 whereas rest also have considered a localization correct if 50% overlap is achieved.

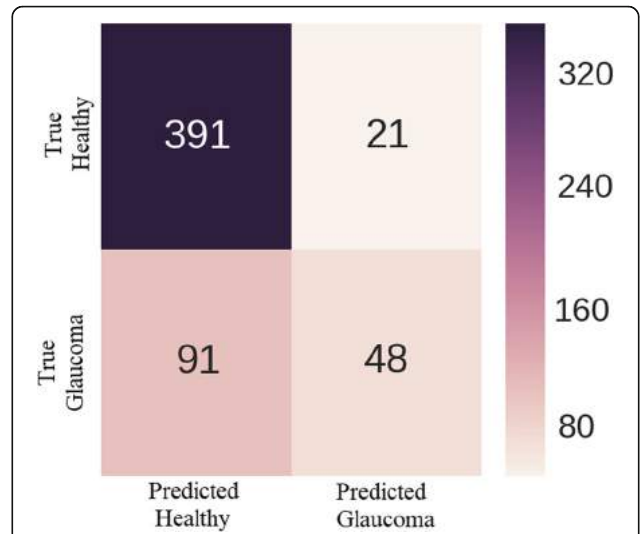

**Fig. 9** Confusion Matrix showing distribution of True Positives, False Positives, and False Negatives

**Table 6** Comparison of obtained AUC with existing state-of-the-art methods using random training

| Performance Metric | [5] | [6] | [38] | [37] | [43] | Our Method |
|---|---|---|---|---|---|---|
| AUC | 0.831 | 0.838 | 0.838 | 0.823 | 0.851 | 0.868 |

As can be seen from Table 4, our automated method performed significantly better than existing heuristics methods, which means that it was able to learn discriminative representation of OD. It should be noted here that heuristics methods are normally designed with a particular dataset in focus. Figure 5 and Fig. 8 show that there exists substantial variations in colour, brightness, contrast and resolution etc. among images of different datasets. Our proposed fully automated method was not trained on any of the four datasets listed in Table 4 and yet it performed superior than those methods tailored specifically for those individual datasets. The average overlap of predicted and actual OD bounding boxes is 84.65% for DIARETDB1, 84.13% for DRIVE, 80.46% for DRIONS-DB and 84.82% for MESSIDOR.

## Results of classification

Due to class imbalance in ORIGA dataset, as shown in Table 1, a stratified sampling technique is implemented where it is made sure that each batch for training contains some of the glaucoma images. This technique is used to prevent any bias towards healthy class. Furthermore, a constant learning rate of 0.0001 along with Adam optimizer and Cross Entropy loss was used during training.

### Results with random training

As no standard split of train and test set is available for this dataset, to compare our model with other recently reported works we first used the same training setup used by most of them [5, 6, 37]. We trained our model repeatedly every time randomly taking 99 images for training and rest for testing. From more than 1500 training runs the best combination of train and test split resulted in overall classification accuracy of 79.67%. For classification of unbalanced datasets, like ORIGA, where number of images in both classes are greatly disproportionate, as evident from Table 1, accuracy alone does not portray true performance of classifier. Therefore, other performance metrics like precision, recall and F1-score are also calculated. Precision indicates the ratio of correct predictions among all the predictions made by any classifier for a certain class. Recall or sensitivity, on the other hand, refers to the fraction of correct predictions for a class out of actual total number of samples in that class. F-score is the harmonic mean of both precision and recall and gives a unified metric to assess classifier's performance. Mathematical definitions of all these performance measures are given below.

**Table 7** Detailed performance measure of CNN classifier using cross validation

| Class | Precision (%) | Recall (%) | F1-Score |
|---|---|---|---|
| Healthy | 82.31 ± 2.88 | 91.86 ± 2.29 | 0.8681 ± 0.246 |
| Glaucoma | 65.52 ± 6.65 | 43.66 ± 4.95 | 0.5231 ± 0.534 |
| Total | 77.97 ± 3.78 | 79.38 ± 3.42 | 0.7788 ± 0.366 |

$$Precision = \frac{TruePositives}{TruePositives + FalsePositives} \tag{1}$$

$$Recall = \frac{TruePositives}{TruePositives + FalseNegatives} \tag{2}$$

$$F1-Score = 2\frac{Precision \times Recall}{Precision + Recall} \tag{3}$$

Class based average precision, recall and F1 scores are tabulated in the Table 5 below.

Figure 9 shows the confusion matrix. It can be observed from the figure that out of 412 healthy images 391 are correctly classified and 21 healthy images are misclassified as having glaucoma. On the other hand, only 48 of total 139 glaucomatous images are correctly classified and 91 images with glaucoma are incorrectly classified as healthy.

Receiver Operating Characteristic (ROC) curve is a popular performance metric used to evaluate the discriminative ability of binary classifier. It uses a varying threshold, on the confidence of an instance being positive, to measure the performance of the classifier by plotting recall (sensitivity) against specificity. Specificity or True Negative Rate (TNR) is defined as,

$$Specificity = \frac{TrueNegatives}{TrueNegatives + FalsePositives} \tag{4}$$

The AUC of this ROC gives a quantitative measure to compare the performance of different classifiers. Table 6 shows the superiority of our model over other comparative studies in terms of AUC.

As most of the works cited in Table 6 reported only AUC as performance metric for their classifiers, we found that for some combinations of 99 train and 551 test images our model was able to achieve higher AUC, 84.87%, than four results in [5, 6, 37, 38] while predicting only healthy class for every test image producing healthy class recall of 1 and glaucoma class recall of 0. It means that the trade-off between sensitivity and specificity of the models can result in higher AUC without learning anything. Therefore, in the absence of clearly defined train and test split and without knowing proportion of healthy and glaucomatous images in both sets, AUC only may not depict the complete and true picture of a classifier. Other performance measures like precision, recall, and f-scores should also be reported for a fair analysis and comprehensive comparison with other models. In case of well-defined train and test split, however, AUC alone might be enough to quantify the classification ability of a model.

**Table 8** Comparison of obtained AUC with existing state-of-the-art methods using cross validation

| Performance Metric | Chen et al. | | Cheng et al. [38] | Xu et al. [37] | Fu et al. [43] | Proposed Model | |
|---|---|---|---|---|---|---|---|
| | [5] | [6] | | | | Random Training | Cross Validation |
| AUC | 0.831 | 0.838 | 0.838 | 0.823 | 0.851 | 0.868 | 0.874 |
| Sensitivity[a] (%) | N/A | N/A | N/A | 58 | N/A | 71 | 71.17 |

[a]The sensitivity is calculated at observed specificity of 85% as done by Xu et al.

### Results with cross validation

Realizing this pitfall in performance evaluation of classifiers, and to facilitate future researchers in thorough comparison of their models we performed 10-fold cross validation on the dataset. The whole dataset was randomly divided into 10 equal portions. In one training session, for example, first part is reserved for testing and other nine are used for training. In next session, second part, for example, is kept aside for testing and rest of nine are used for training. Average is taken over 10 training sessions and the accumulative test accuracy is found to be 79.39% ± 3.42%. Class based precision recall and f1 score are tabulated in Table 7.

The comparison of AUC obtained using cross validation with other works is summarized in Table 8 which clearly shows that the proposed network outperforms stat-of-the-art results for glaucoma classification on ORIGA dataset. Figure 10 shows sample images of correctly and incorrectly classified glaucoma and healthy images.

Data augmentation was also performed to study its effects on the accuracy of classification. We performed horizontal and vertical flips and cropping 227 × 227 × 3 patches from four corners and centre of 256 × 256 × 3 extracted images of OD. However, the experiments performed with and without data augmentation showed no significant difference between the performances of both approaches. We also explored the effect of network complexity on the classification accuracy. For this purpose, we used Alexnet [59] as the reference model and assessed the impact of the number of layers on the network's performance given all the other conditions are the same. It was observed that increasing network complexity actually deteriorated the accuracy of the classifier. The reason for this performance degradation can be small size of the dataset. Deeper network shave a habit of overfitting during training when not enough training samples are provided. The networks working better than others had four convolutional layers. The best working model among all the different versions tried is used for classification and is the one shown in Fig. 7.

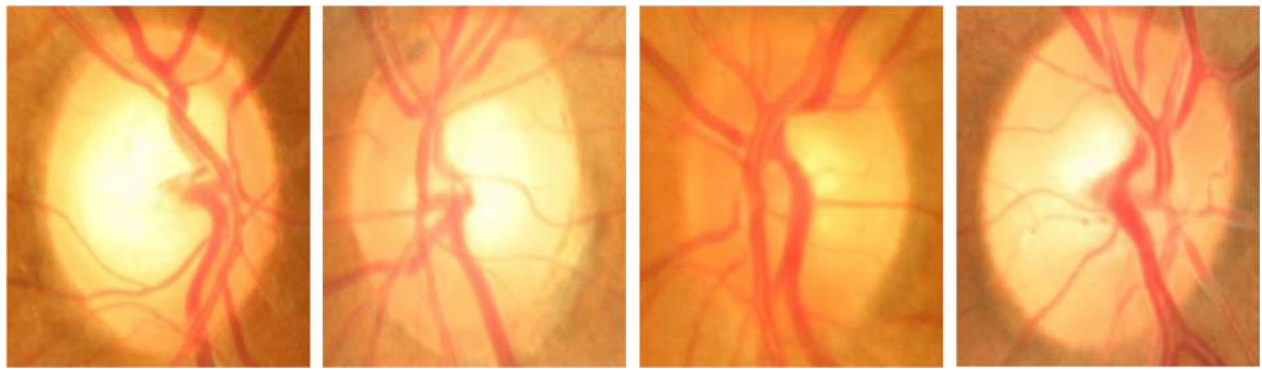

**Fig. 10** Some examples of correct and incorrect glaucoma classification using DCNN. **a** Glaucoma correctly classified **b** Glaucoma incorrectly classified **c** Healthy image correctly classified **d** Healthy image incorrectly classified

## Conclusion

We proposed a two-stage solution for OD localization and classification for glaucoma screening. Realizing that OD is instrumental in examining the eye for various diseases including glaucoma, we first employed a fully automated disc localization method based on faster RCNN. This method completely eliminates the need for development of dataset-specific empirical or heuristic localization methods by providing robust and accurate localization across a broad spectrum of datasets. The performance of this fully automated systems sets new state-of-the-art results in six out of seven publicly available datasets with IOU greater than 50%.

The classification of images into diseased and healthy using CNN has also been investigated. Although some researchers have reported around 95% accuracy on private datasets or carefully selected small set of images from public datasets, the classification accuracy and ROC AUC for publicly available ORIGA dataset has been challenging to improve. In spite of the fact that we were able to achieve significantly higher AUC on ORIGA with both random training and *k*-fold cross validation, the detailed performance measures of the classifier reveal that the network has difficulty in learning discriminative features to classify glaucomatous images in this public dataset. It appears that the fine grained discriminative details in the images of this dataset are lost with the increase in the hierarchy of the network. Therefore, more effort is required to tailor some classifiers capable of identifying glaucomatous images with reliability.

### Abbreviations
AUC: Area under the curve; CAD: Computer-aided diagnosis; CDR: Cup-to-disc ratio; CNN: Convolutional neural network; DCNN: Deep CNN; GT: Ground truth; HT: Hough transform; ICDR: International Clinical Diabetic Retinopathy; IOP: IntraOcular pressure; IOU: Intersection over union; ISNT: Inferior superior nasal temporal; LRN: Local response normalization; OC: Optic cup; OCT: Optical coherence tomography; OD: Optic disc; ONH: Optic nerve head; PPA: Peripapillary atrophy; RCNN: Regions with CNN; ReLU: Rectified Linear Unit; RGB: Red Green Blue; ROC: Receiver operating characteristic; ROI: Region of interest; RPN: Region proposal network; SBF: Sliding band filter; SSIM: Structured SIMilarity; TNR: True negative rate; WHO: World Health Organization

### Acknowledgements
Not Applicable.

### Authors' contributions
MNB designed and implemented the methodology and wrote the manuscript. MIM and SA conceptualised the project, administered the experiments, and revised the draft through several iterations. SAS helped in design and implementation of the proposed method. AD and FS supervised the overall project thereby imparting their inputs at different levels of work execution. WN provided ophthalmological insight and validated ground truth. All authors approved the final manuscript.

### Funding
Muhammad Naseer Bajwa is supported by National University of Science and Technology (NUST), Pakistan and Higher Education Commission (HEC) of Pakistan for PhD funding under 'Prime Minister Program for Development of PhDs in Science and Technology'.

### Availability of data and materials
All databases used in this work are publically available.
*ORIGA:* Zhang, Z., Yin, F.S., Liu, J., Wong, W.K, Tan, N.M., Lee, B.H., Cheng, J., Wong, T.Y.: ORIGA-light: An online retinal fundus image database for glaucoma analysis and research. In: Engineering in Medicine and Biology Society (EMBC), 2010 Annual International Conference of the IEEE, pp. 3065–3068 (2010). IEEE.
*HRF Image Database:* Budai, A., Bock, R., Maier, A., Hornegger, J., Michelson, G.: Robust vessel segmentation in fundus images. International journal of biomedical imaging (2013).
*OCT & CFI:* Mahmudi, T., Kafieh, R., Rabbani, H., Akhlagi, M., et al.: Comparison of macular OCTs in right and left eyes of normal people. In: Medical Imaging 2014: Biomedical Applications in Molecular, Structural, and Functional Imaging, vol. 9038, p. 90381 (2014). International Society for Optics and Photonics.
*DIARETDB1:* Kauppi, T., Kalesnykiene, V., Kamarainen, J.-K., Lensu, L., Sorri, I. Raninen, A., Voutilainen, R., Kalviainen, H., Pietila, J.: DIARETDB1 diabetic retinopathy database and evaluation protocol. In: Medical Image Understanding and Analysis, 2007, Proceedings of 11th Conference On (2007).
*DRIVE:* Staal, J.J., Abramoff, M.D., Niemeijer, M., Viergever, M.A., van Ginneken, B.: Ridge based vessel segmentation in color images of the retina. IEEE Transactions on Medical Imaging 23(4), 501–509 (2004).
*DRIONS-DB:* Carmona, E.J., Rincón, M., García-Feijoó, J., Martínez-de-la-Casa, J.M.: Identification of the optic nerve head with genetic algorithms. Artificial Intelligence in Medicine 43(3), 243–259 (2008).
*Messidor:* Decencire, E., Zhang, X., Cazuguel, G., Lay, B., Cochener, B., Trone,C., Gain, P., Ordonez, R, Massin, P., Erginay, A., Charton, B., Klein, J.-C.: Feedback on a publically distributed image database: The messidor database. Image Analysis & Stereology 33(3), 231–234(2014). doi:https://doi.org/10.5566/ias. 1155

### Ethics approval and consent to participate
Not Applicable.

### Consent for publication
Not Applicable.

### Competing interests
The authors declare that they have no competing interests.

### Author details
[1]Fachbereich Informatik, Technische Universität Kaiserslautern, 67663 Kaiserslautern, Germany. [2]Deutsche Forschungszentrum für Künstlicheintelligenz GmbH (DFKI), 67663 Kaiserslautern, Germany. [3]Deep Learning Laboratory, National Center of Artificial Intelligence, Islamabad 46000, Pakistan. [4]School of Electrical Engineering and Computer Science (SEECS), National University of Sciences and Technology, H-12, Islamabad 46000, Pakistan. [5]Ophthalmology Clinic, Rittersberg 9, 67657 Kaiserslautern, Germany.

## Publisher's Note